\documentclass[letterpaper, 10 pt, conference]{ieeeconf}  % Comment this line out if you need a4paper

\IEEEoverridecommandlockouts                              % This command is only needed if 
\usepackage{graphicx} % for pdf, bitmapped graphics files
\usepackage{subcaption}
\usepackage[normalem]{ulem}
\usepackage{tikz}

\title{\LARGE \bf
    When May I Help You? On The Effect of Proactivity on Group Human-Robot Collaboration
}
\author{Thomas Vitry$^{1,2}$, Vanessa Maeder$^{1}$, Kieran von Valeburg$^{1}$, Asihati Hazaiti$^{1}$, Doga Deniz Ates$^{1}$
\\Connor Gäde$^{1}$, Jan-Gerrit Habekost$^{1}$, Dennis Becker$^{1}$ and Stefan Wermter$^{1}$% <-this % stops a space
\thanks{The authors gratefully acknowledge support from the German Research Foundation DFG under project LUMO (No 551629603) and funding from Horizon
Europe under the MSCA grant agreement No 101168792 (SWEET), No 101072488 (TRAIL), and No 101226624  (GREET).}
\thanks{$^{1}$Knowledge Technology, University of Hamburg, Hamburg, Germany. Emails: thomas.vitry@studium.uni-hamburg.de, \{stefan.wermter, dennis.becker-1\}@uni-hamburg.de} %
\thanks{$^{2}$Ecole Normale Superieure de Rennes, Rennes, France}%
}

\begin{document}

\maketitle
\thispagestyle{empty}
\pagestyle{empty}

%%%%%%%%%%%%%%%%%%%%%%%%%%%%%%%%%%%%%%%%%%%%%%%%%%%%%%%%%%%%%%%%%%%%%%%%%%%%%%%%
\begin{abstract}

    Robot initiative is a central challenge in multi-party human-robot collaboration. A robot that contributes without being addressed may provide timely support, but it may also disrupt coordination, divide attention, or interrupt turn-taking; a robot that waits to be addressed may preserve human control, but it may also miss opportunities to assist. We investigate this design challenge in a collaborative escape room in which pairs of participants work with a humanoid robot under either a reactive interaction model, where the robot responds only when addressed, or a proactive model, where it listens continuously, contributes autonomously, and periodically re-initiates interaction. We evaluate both models using puzzle-solving performance, interaction frequency, and participant ratings on the Godspeed and RoSAS scales. The proactive model substantially increases interaction frequency, whereas the reactive model shows a descriptively higher overall success rate (92.86\% vs. 71.42\%). The strongest differences emerge when prior experience and personality are taken into account: participants with LLM experience solve the early puzzles faster in the reactive condition, and participants with prior robot experience show modified evaluations of proactive and reactive interaction as do introverted participants. These findings demonstrate that the effects of robot initiative are simultaneously shaped by users’ prior experience, personality traits and more generally by the needs of the group.

\end{abstract}

% \copyrightnotice

%%%%%%%%%%%%%%%%%%%%%%%%%%%%%%%%%%%%%%%%%%%%%%%%%%%%%%%%%%%%%%%%%%%%%%%%%%%%%%%%
\section{Introduction}

    As robots take on more collaborative roles, their contributions to an interaction depend on what they do, when they act, and how they act~\cite{SurveyHRIGoodrich, HoffmanCollaboration}. In shared activities, successful human-robot interaction depends on both joint task completion and the robot contributing at appropriate moments without disrupting coordination between partners. These challenges become especially important in multi-party settings, where a robot must balance helpfulness with the risk of interrupting or distracting human collaborators~\cite{HoffmanCollaboration, nigro2024socialgrouphumanrobotinteraction}.

    These coordination demands make robot initiative a central challenge in collaborative human-robot interaction~\cite{SurveyHRIGoodrich, HoffmanCollaboration}. A robot that waits to be addressed may preserve human control, but it may also miss opportunities to support the task. A robot that contributes on its own may help maintain progress, but it may also interrupt, overstep, or change how the interaction is experienced~\cite{zhang2015human, nikolaidis2017human, JamshadProactiveRobotInTeam}. This tension becomes even more apparent for language-enabled robots, whose contributions are conveyed not only through physical action but also through language and dialogue, making them more visible as both task support and social behavior~\cite{BeckerAttitude, Multi-party, ahn2022icanisay}.
    
    The central research question in this work is how a robot should participate in multi-party collaboration: by following a reactive interaction style and waiting to be addressed, or by taking a more proactive role and contributing on its own. This question is investigated in a collaborative escape room scenario in which pairs of participants interact with a humanoid robot assistant under time pressure and a shared goal, creating repeated opportunities for the robot either to wait for explicit requests or to contribute autonomously. Specifically, this study compares a reactive interaction model, in which the robot responds only when explicitly addressed, with a proactive model, in which it continuously listens, contributes more autonomously, and periodically re-initiates interaction. The study assesses how these interaction styles affect task performance, interaction behavior, and social perception of the robot. In doing so, it provides an empirical comparison of proactive and reactive interaction styles for a language-enabled robot in a multi-party cooperative human-robot setting and introduces a collaborative escape room setup for examining the effects of robot initiative in an interaction involving a human pair and a robot.

    % Recommended at the end of the introduction to also add results here, mentioning specific values and being more explicit with measurements
    
%%%%%%%%%%%%%%%%%%%%%%%%%%%%%%%%%%%%%%%%%%%%%%%%%%%%%%%%%%%%%%%%%%%%%%%%%%%%%%%%
\section{Background}

    Human-robot interaction research increasingly examines robots as tools and collaborative partners whose timing, communication, and social behavior shape the quality of joint activities. In shared tasks, interaction quality depends on more than task completion alone; perceived fluency, coordination, and the robot's ability to participate appropriately in a multi-person interaction are also central to user experience~\cite{SurveyHRIGoodrich, HoffmanCollaboration, nigro2024socialgrouphumanrobotinteraction}. The interaction model, therefore becomes a key design variable in collaborative scenarios, especially when robots must balance helpful initiative with respect for human control.

    In addition to behavior, the social perception of these robot partners is also affected by the Uncanny Valley. According to this theory, the more a robot resembles a human, the more people feel warmer toward it; however, when the resemblance reaches a very high level, the imperfect small details of the robot create a feeling of discomfort or unease in people \cite{uncanny1}. While it was previously thought that there was only one valley, new research shows that there may actually be two uncanny valleys. In particular, users can feel uncomfortable even at much lower levels of similarity when the robot's appearance and social behavior do not match \cite{uncanny2}.

    A main aspect of robot behavior in collaborative interaction is whether the robot takes initiative proactively or responds reactively. Prior work suggests that proactive support can assist task progression by anticipating user needs and taking initiative, but its effects are not uniformly positive~\cite{nikolaidis2017human, jamshad2026human, gao2019fast}. Proactive behavior can improve coordination and reduce idle time. However, it also alters expectations regarding autonomy, assistance, and the division of control~\cite{zhang2015human, JamshadProactiveRobotInTeam}. Its value therefore depends on the task context and users interpretation of it, rather than being universally beneficial~\cite{HoffmanCollaboration,zhang2015human,JamshadProactiveRobotInTeam}. This is relevant in collaborative team-based tasks, where timely support should not undermine the human partner's sense of agency.

    Further work shows that individuals evaluate robots both socially and instrumentally. Social categorization can already shape these perceptions, as robots framed as in-group members are anthropomorphized more strongly and evaluated more positively~\cite{KuchenbrandtGroupMembership}. Robot personality, gaze, facial expression, and conversation style further shape perceived likability, trust, engagement, and performance in interactions~\cite{DryerExtroversionIntroversionHRI, BeckerAttitude, NgRoleplaying}. Personality cues can affect preference and trust, while gaze and facial expressions influence collaboration and perceived intelligence~\cite{personalityandriella, LimRobotPersonality, FuExpressions}. With language-enabled robots, these distinctions become even more important, as robot behavior is expressed through action timing and also through language, dialogue structure, and, in some systems, personality-consistent responses~\cite{gade2024elmira, BeckerAttitude, Multi-party}. In parallel, LLMs can be grounded in robot capabilities and executable skills, enabling long-horizon embodied action from natural-language instructions~\cite{ahn2022icanisay}. These developments suggest that robots are increasingly experienced both as physical team members and as language-enabled social partners.

    Game-based and escape room-like settings provide a useful testbed for studying human-robot interaction under time pressure, coordination demands, and dialogue-rich collaboration. They have been used to examine engagement and collaborative dynamics in structured interactive tasks~\cite{dong2025happiness, NgRoleplaying, buchem2025escape}. Prior research in these settings also examines trust, emotion, and interactional cues, including affective influences on performance and the effects of robot facial expressions and gaze in collaborative interaction~\cite{gao2019fast, FuExpressions}. However, much of this work tends to emphasize either performance-related outcomes, including game progress and task success~\cite{dong2025happiness, buchem2025escape}, or perceptual and evaluative factors, such as trust, affect, and social interaction cues~\cite{gao2019fast, dong2025happiness, FuExpressions}. Comparatively less attention is given to how proactive and reactive interaction styles shape multiple aspects of the interaction, including task progress and completion, the number of participant-initiated exchanges, and subjective evaluations of the robot and the overall collaboration experience. This broader perspective is important because the same robot may be experienced simultaneously as a team member in the task, a social actor in the interaction, and a conversational interface.
    
    To capture differences in how the robot is perceived, human-robot interaction research commonly relies on standardized perception measures. The Godspeed Questionnaire Series~\cite{BartneckGodspeed, BartneckGodspeedReference} remains widely used to assess anthropomorphism, animacy, likability, perceived intelligence, and perceived safety. The Robotic Social Attributes Scale (RoSAS)~\cite{CarpinellaRoSAS} complements these dimensions by focusing more explicitly on social evaluations of the robot, particularly warmth, competence, and discomfort~\cite{Pan2018EvaluatingSP, NeuenswanderShortRoSAS}. Methodological work also highlights limitations of relying on Godspeed alone, including strong inter-correlations among several of its indices, which motivates combining it with complementary measures~\cite{HoAlternativeToGodspeed}. In our study, combining Godspeed and RoSAS allows us to assess the broad perceptions of the robot, and the more specific social evaluations that may vary between proactive and reactive interaction styles.

%%%%%%%%%%%%%%%%%%%%%%%%%%%%%%%%%%%%%%%%%%%%%%%%%%%%%%%%%%%%%%%%%%%%%%%%%%%%%%%%
\section{Methods}

\subsection{Experiment}
\label{sec:experiment}

% We aim at studying the preferred interaction model of a robot when placed in a social human setting. To that end, we establish a number of requirements. We want the setting to feature multiple humans, cooperatively working to solve on or more problems. To avoid influencing the perception of the subjects, the robot should be presented neutrally as a potential assistance in the problem solving all the while not a necessity, making any interaction with the robot voluntary and not forced by the scenario.

\subsubsection{Study Design}
\begin{figure}
  \vspace{5pt}
  \includegraphics[width=\linewidth]{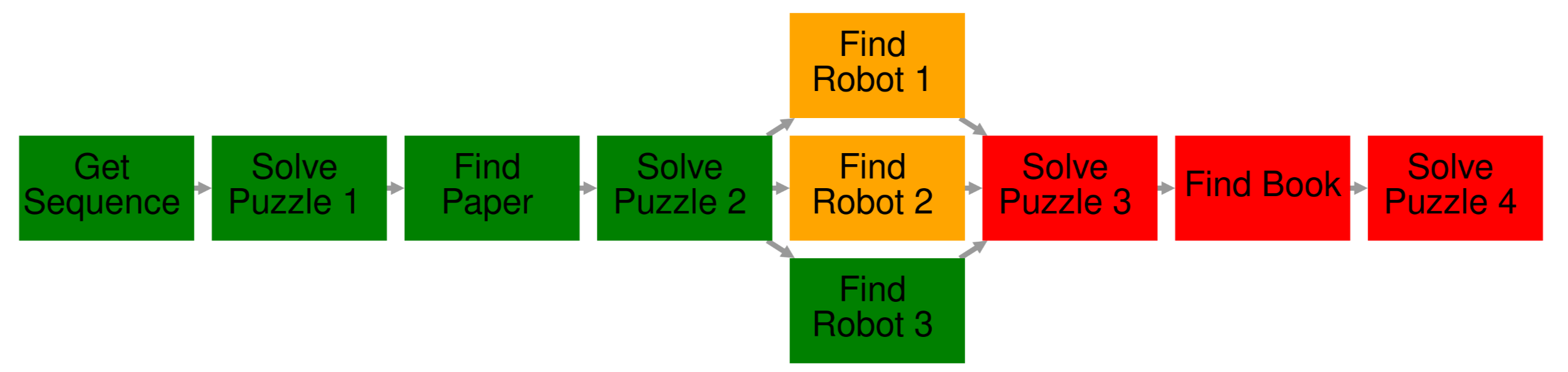}
  \caption{Puzzle graph of the escape room. Nodes represent sub-puzzles, and edges represent their requirements. Green nodes are solved, orange nodes are solvable, and red nodes are not yet solvable.}
  \label{fig:puzzletree}
\end{figure}

To study the effect of the robot's proactivity model in multi-human social settings, we design an escape room experiment for two participants. Participants are introduced to the escape room scenario and then placed in a room containing various objects and pieces of furniture. Within a limited time, they have to solve a series of puzzles distributed throughout the environment. This setting simulates a natural social context in which participants can interact freely with both the robot and each other while working toward a shared goal. To encourage natural interaction with the robot, it is introduced as an additional participant that, within the escape room scenario, has access to additional knowledge about the environment. Progress through the escape room is tracked by the experimenters using a directional graph of the sub-puzzles (see Figure~\ref{fig:puzzletree}). To prevent participants from becoming stuck, hints are periodically provided by an automatic scheduler. In exceptional cases, experimenters can also provide manual hints once no further automatic hints are available for the current state of the experiment.

% \begin{figure}
%   \includegraphics[width=\linewidth]{src/problem_tree_flow_v2.png}
%   \caption{Puzzle graph of the escape room. Nodes represent sub-puzzles, and edges represent their requirements.} %Green nodes are solved, orange nodes are solvable, and red nodes are not yet solvable.}
%   \label{fig:puzzletree}
% \end{figure}

\subsubsection{The Robot}
The robot used in this study is the NICO (Neuro-Inspired COmpanion) robot~\cite{NICO}, a child-sized humanoid robot with an expressive LED face and anthropomorphic hands. The robot embodies a ChatGPT 4o LLM through a modified version of the ELMiRA~\cite{gade2024elmira} framework. It continuously captures participants' speech using individual wearable microphones. The captured audio is transcribed into English sentences using a Whisper model~\cite{radford2023robust}. These sentences are then passed to the LLM together with information about the identity of the speaker to form its input. The LLM then produces a list of actions for the robot to perform. Available actions include speaking, displaying a specific emotion, pointing, and grasping objects. The robot also has actions allowing him to gather additional information. It can use its cameras to take a picture of the environment and can query its memory about a specific object. Both of these actions produce additional input that is included in the next call to the LLM. The \textit{memory system} is implemented as a dictionary containing object- and concept-specific information about the room, stored under specific keywords. When the robot queries the memory system, it submits a keyword to be searched for. Embeddings of the query and the stored keywords are computed and compared before the memory system returns the memory associated with the most similar keyword. If the keyword similarity is too low, the memory system instead returns that no relevant memory is found for the query.

Through this memory system, the robot has on-demand access to environmental information with varying levels of usefulness. It is however uninformed as to the escape room itself and has no innate knowledge of the puzzles, their progression, or their intended solutions. For example, the robot may know the name of a painting in the room and some related facts, but it has no built-in knowledge of whether that painting is part of a puzzle. It can, however, still infer a possible connection if prompted to retrieve information about the painting and given relevant details about the puzzle context. When provided with sufficient descriptions, the robot is even able to solve puzzles by itself using its deductive abilities. Both memories and hints are provided directly to the robot as additional inputs to the LLM, along with the corresponding source tag. The robot then transmits this information to the participants according to its initial prompt.

\subsubsection{Interaction Models}
We implement two separate interaction models, which are used interchangeably across experiments to compare the effect of robot proactivity. The first, referred to as the reactive model, is designed to function similarly to a speech assistant. It responds only to utterances containing its name and otherwise speaks only to transmit hints. The second, referred to as the proactive model, listens and responds to everything said in the room, as if it were being continuously addressed. In addition, if no interaction occurs for 90 seconds, the proactive model autonomously addresses the participants to ask about their current progress and whether it can help.

\subsubsection{Outline of the Escape Room}
\label{seq:experimental_setup}

\begin{figure*}[t]
    \vspace{5pt}
    \centering
    \includegraphics[width=0.99\linewidth]{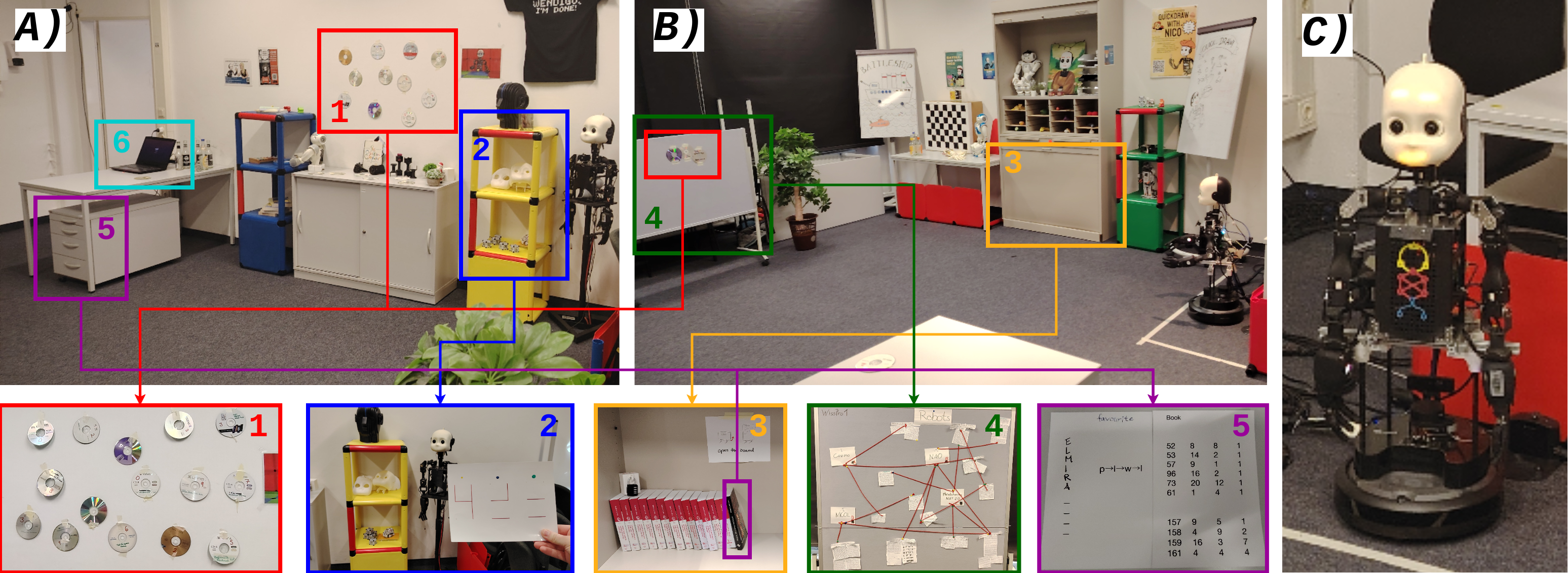}

    \caption{Overview of the escape room from \textit{A)} the back corner and from \textit{B)} the participant entrance. In Puzzle one (red), participants arrange CDs distributed across the room (1) into the robot’s startup sequence, as hinted by inscriptions on the CDs from the board (4). The sequence can be recovered from the robot memory and is also provided automatically after $420$ seconds. This puzzle introduces the robot as a source of scenario-specific knowledge. In Puzzle two (blue), participants match furniture patterns in the room to a reference sheet (2). Robot use is not required beyond optional hints, making this puzzle a baseline for interaction when assistance is unnecessary. This puzzle leads through the shelf (3) to the opening of the board (4). In Puzzle three (green), participants identify robots placed around the room by name, either by asking NICO or by using written descriptions from the board (4). This puzzle assesses willingness to rely on the robot when alternative information sources are available. In Puzzle four (purple), the password is encoded in a book using a page-line-word-letter scheme (5). Participants are expected to identify the correct book by asking for the supervisor’s favorite book, although the answer is also sent automatically after $640$ seconds and can also be inferred from the book’s placement and singular topic. The final password is then entered on the computer (6), also used through the experiment to display the remaining time.
    The NICO robot \textit{C)} is located in a reserved area (right of \textit{B)}).
    % Setup 1 is the task of finding digits to unlock a 3-digit lock, where the digits are obtained by complementing numbers based on the color of the paper and the shelf. Setup 2 is the task of solving a 5-digit lock from a set of programmable discs that have numbers associated with NICO's startup sequence, which can be learned by asking NICO itself. 
    % Setup 3 is about solving a 4-digit lock using a board full of 4 robotics papers describing their characteristics, where the corresponding robots are physically present in the room and labeled with numbers. The final task, as shown in sub-figures 4 and 5, involves finding the supervisor’s favorite book. An information sheet reveals the final password by highlighting the password's letters in the book using a number sequence consisting of page, line, word, and letter numbers.
    }
   \label{fig:escape_room}
\end{figure*}

The escape room scenario places participants in the role of interns whose supervisor has left for a conference. Having forgotten to submit their presentation before leaving, the participants are placed in the supervisor's office and tasked with finding, with the help of the supervisor's robot assistant NICO, the password needed to unlock the computer and submit the presentation within a 60-minute deadline. The escape room is designed as a linear sequence of four puzzles. An overview of the escape room is shown in Figure~\ref{fig:escape_room}.

\subsubsection{Hypothesis}
\label{seq:hypothesis}

Based on the characteristics of the two interaction models, we formulate the following hypothesis: \textit{1)} The proactive interaction model leads to more interaction than the reactive one. 
\textit{2)} Due to the higher possibility of a proactive robot being perceived as misleading or leading to confusion ~\cite{jamshad2026human}, more extreme outcomes are expected for the proactive robot.
%\textit{2)} Because of the agreeable nature of LLM, the proactive model leads to more inconsistent and extreme outcome than the reactive one.
\textit{3)} Effects of proactive and reactive models are dependent on prior experiences ~\cite{TrustPriorExperience2017} and personality factors \cite{PersonalityHRI2017} such as escape room experience, LLM experience, robot experience and extraversion.
%Options for H3: 
%1: Effects of proactive and reactive models are dependent on prior experiences and personality of the participants.
%2: Effects of proactive and reactive models are dependent on prior experiences and extraversion of the participants.
%3: Effects of proactive and reactive models are dependent on prior experiences \cite{TrustPriorExperience2017} and the personality of the participants \cite{PersonalityHRI2017}.
%4: Effects of proactive and reactive models are dependent on four confounding variables: escape room experience, LLM experience, robot experience \cite{TrustPriorExperience2017} and extraversion \cite{PersonalityHRI2017}.
%5: Effects of proactive and reactive models are dependent on prior experiences \cite{TrustPriorExperience2017} and the extraversion of participants \cite{PersonalityHRI2017}.
%6: Effects of proactive and reactive models are dependent on prior experiences \cite{TrustPriorExperience2017} and the personality of participants with an emphasis on the extraversion \cite{PersonalityHRI2017}.

\subsection{Experimental Procedure}

\subsubsection{Measures}
\label{sec:measures}

To evaluate this study, we collect both objective performance metrics and subjective measures derived from participants' responses in pre- and post-experiment questionnaires. During the experiment, the time taken to solve each puzzle, the hints provided to participants, and the number of interactions participants initiate with the robot were recorded.

Additional data are obtained through two questionnaires. The pre-experiment questionnaire collects basic demographic information (age and self-reported gender) as well as openness and extraversion personality data using a subset of the BFI-10~\cite{rammstedt2007measuring}. The post-experiment questionnaire measures robot perception using RoSAS and Godspeed, and also gathers information about participants' prior experience with robots, escape rooms and LLMs. Escape room experience is measured as the absolute number of prior escape room participation. Participants are considered experienced if they have previously completed at least three escape rooms. Experience with LLMs and robots is reported on a five-point Likert scale; participants are considered inexperienced if they respond with one or two, and experienced if they respond with three to five. For extraversion, high and low extraversion is used based on the median ($Mdn = 3.0$) of the participants extraversion.

\subsubsection{Conduct of the Experiment}

Before entering the escape room, participants complete the pre-experiment questionnaire to provide personality and demographic data. These results are not used to influence the rest of the experimental setup. Participants are then briefed on the scenario and introduced to the robot, together with a short explanation of how to interact with it according to the interaction model assigned in advance. They are here made aware that the robot works using an LLM. They are not informed that multiple interaction models exist. Participants are then left in the room with the robot, at which point a 45-minute timer begins.

While the experiment is ongoing, progression is continuously monitored by experimenters located outside the room using a surveillance camera and a microphone, although no recordings are made. The experimenters use the puzzle graph to track the flow of the game and provide manual hints when necessary. The room itself is not locked, and participants can stop the experiment at any time upon request. After the 45-minute timer has run out, participants are offered the option to continue for an additional 15 minutes if some puzzles have not yet been fully solved. They can also choose to stop the experiment at that point.

After completing the escape room, regardless of success or failure, participants are asked to complete the post-experiment questionnaire to evaluate their interaction experience with the robot during the study. The experimenters then conduct a final exchange to answer participants' questions about the escape room and the robot.

\subsubsection{Participants}

A total of 60 participants took part in the study. All experiments were conducted in English and carried out in pairs, with the robot serving as an additional interaction partner. After excluding incomplete data, 56 participant records were included in the analysis, corresponding to 28 pairs. Condition assignment was performed in advance at the pair level, resulting in 14 pairs in the proactive condition and 14 pairs in the reactive condition.

In the recruited sample, 18 participants self-identified as women, 37 as men, 2 as diverse, and 3 preferred not to disclose their gender. Participants ranged in age from 19 to 55 years ($M = 28.07, SD = 14.40$). All participants provided informed consent prior to the study. The experiment was conducted in accordance with the Ethics Commission of the Department of Informatics at the University of Hamburg.

% Many participants came from a computer science background; 20 participants reported prior experience with robots and 42 reported prior experience with LLMs.

%%%%%%%%%%%%%%%%%%%%%%%%%%%%%%%%%%%%%%%%%%%%%%%%%%%%%%%%%%%%%%%%%%%%%%%%%%%%%%%%
\section{Results}

We first assess the robot perception measures and the objective performance measures described in Section \ref{sec:measures} using Welch’s t-tests. As expected our first hypothesis holds, as there is a significant difference in the number of interactions ($p < .001$) between the reactive interaction model ($M = 29.07, SD = 12.59$) and the proactive interaction model ($M = 97.86, SD = 21.35$). However, there are no significant differences between the interaction models for the robot perception measures from Godspeed and RoSAS or for the objective performance measures. By using Fisher’s exact test a non-significant trend ($p = .077$) toward a lower success rate in the proactive condition ($71.42\%$) than in the reactive condition ($92.86\%$) can be observed.

\begin{figure}
\vspace{5pt}
  \includegraphics[width=0.99\linewidth]{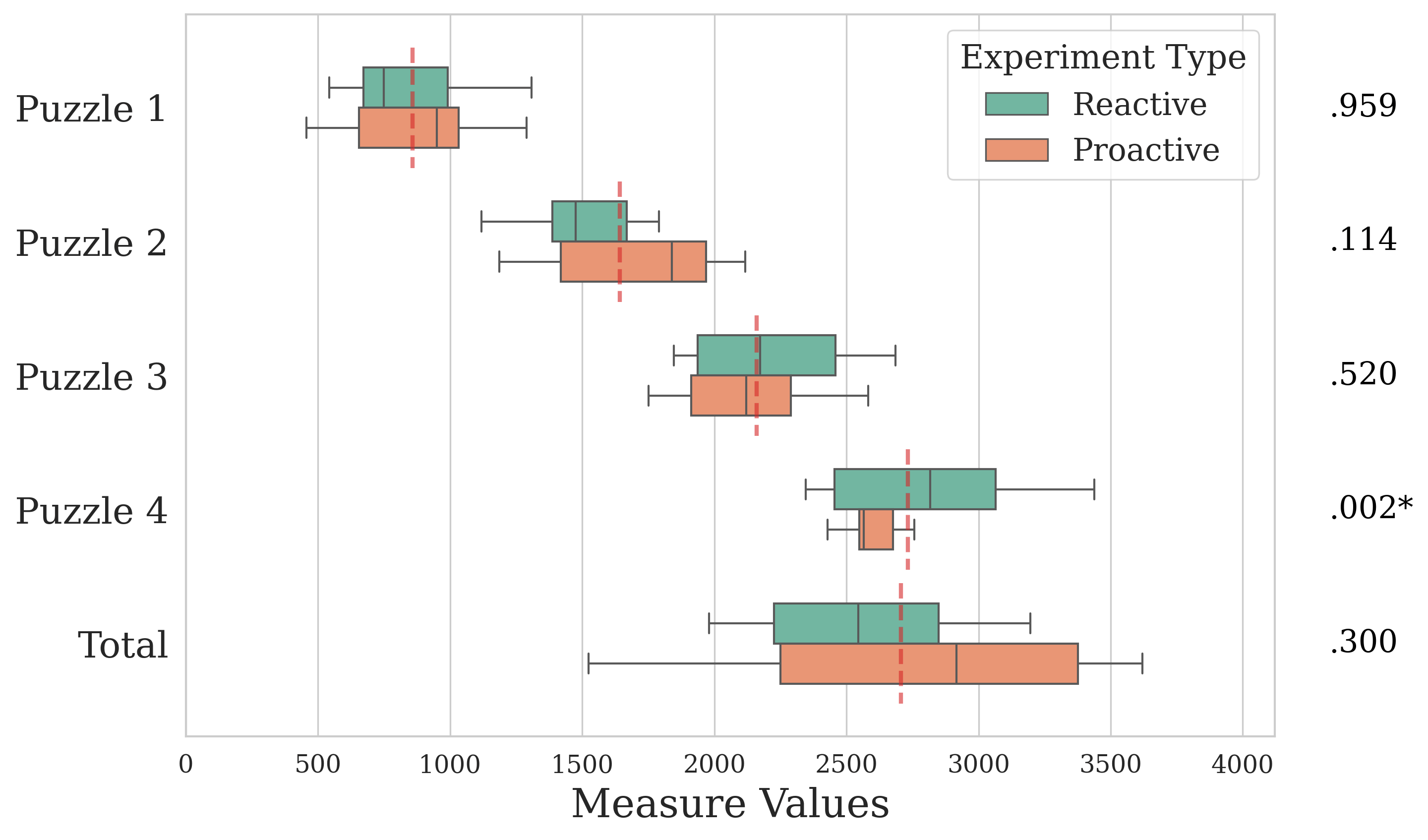}
  %\caption{Timeline of the escape room, with the solving times for each of the puzzles by interaction model visible as box plots as well as the average solving time represented by the red lines. As shown, the reactive interaction model has a lower per puzzle solving time for Puzzle two than the proactive model. It can also be seen that the solve times for the proactive interaction model are more widely spread than those for the reactive model, an effect amplifying with each consecutive puzzle.}
  \caption{Timeline of the escape room, showing puzzle-solving times by interaction model. Box plots indicate the distribution of solving times for each puzzle, and red dashed lines mark the mean solving time. Reported values indicate $p$-values, *$p < .05$, **$p < .001$.}
  \label{fig:Timeline}
\end{figure}

\begin{figure}
\vspace{5pt}
  \includegraphics[width=0.99\linewidth]{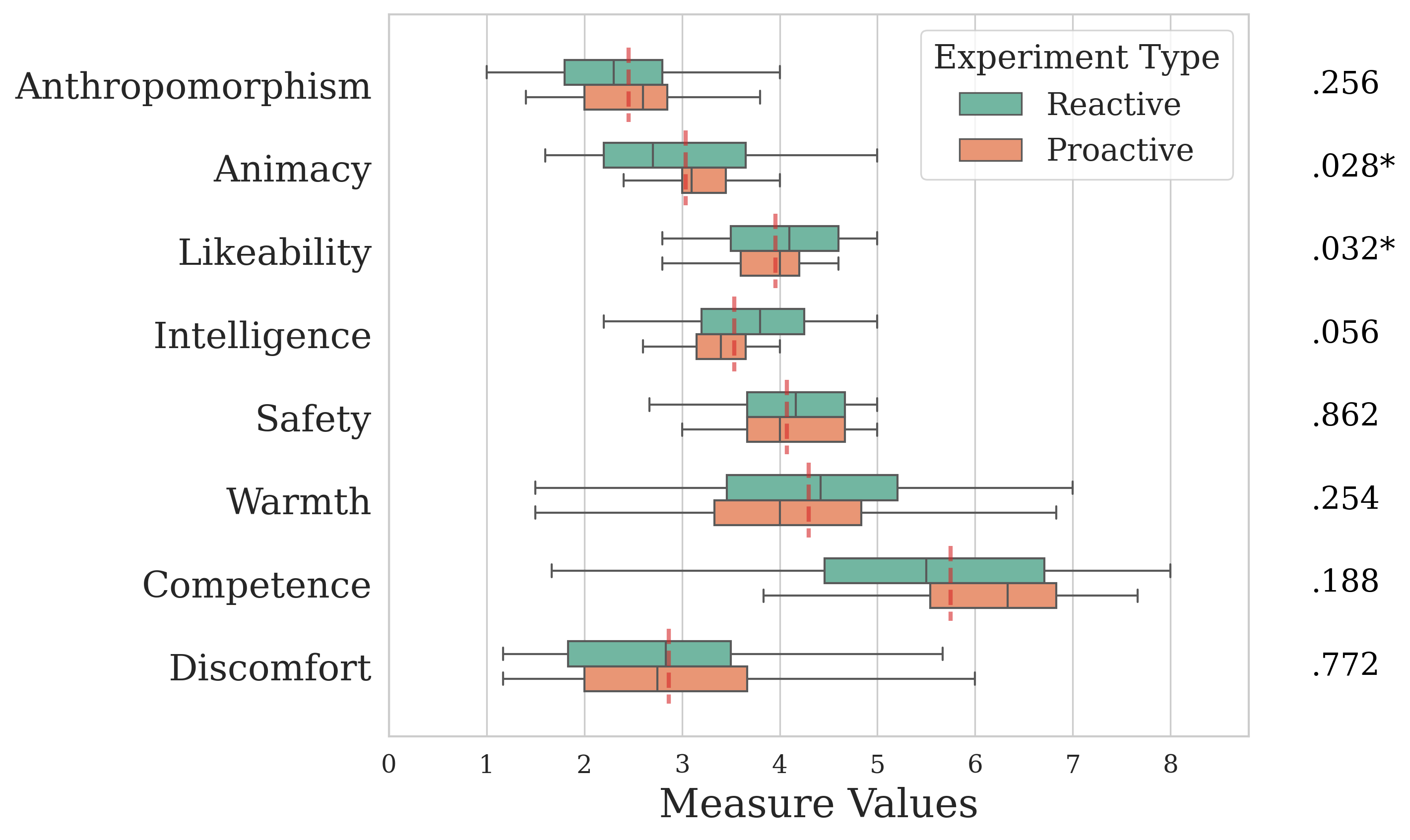}
  %\caption{Timeline of the escape room, with the solving times for each of the puzzles by interaction model visible as box plots as well as the average solving time represented by the red lines. As shown, the reactive interaction model has a lower per puzzle solving time for Puzzle two than the proactive model. It can also be seen that the solve times for the proactive interaction model are more widely spread than those for the reactive model, an effect amplifying with each consecutive puzzle.}
  \caption{Godspeed and RoSAS scores for the proactive and reactive interaction models. Box plots indicate the distribution of each measure for the proactive and reactive models, respectively, and red dashed lines mark the mean for each model. Reported values indicate $p$-values, *$p < .05$, **$p < .001$.}
  \label{fig:perception_variance}
\end{figure}

% reference the HYPOTHESIS
The Fligner–Killeen test is used to test the second hypothesis by assessing the differences in variance between the two interaction models. Significant differences in variance emerge for the time participants need to solve some of the puzzles, as well as for several robot perception measures. As illustrated in Figure \ref{fig:Timeline}, a significant difference in variance is observed for the time required to solve Puzzle four between the reactive model ($M = 654.28, SD = 353.99$) and the proactive model ($M = 463.48, SD = 175.21$). We also identify significant differences in variance for the Godspeed measures Animacy and Likeability, as shown in Figure \ref{fig:perception_variance}. For Animacy, the reactive interaction model shows significantly higher variance ($p = .028; M = 2.86, SD = 0.88$) than the proactive model ($M = 3.21, SD = 0.55$). Similarly, for Likeability, the reactive model ($M = 4.03, SD = 0.72$) displays significantly higher variance ($p = .032$) than the proactive model ($M = 3.89, SD = 0.47$). Across both interaction models, the RoSAS measures Warmth, Competence, and Discomfort also show generally high variance. Thus, despite the proactive interaction model displaying a high variance, said variance is not significantly higher than the variance of the reactive model, refuting our second hypothesis.

To further explore our third hypothesis, we account for four confounding variables: prior escape room, LLM and robot experiences and extraversion. No correlation is found among these experience variables, and we therefore treat them as independent within the context of this study. All conditioned results are assessed using Welch’s t-tests and effect sizes computed with Cohen’s d if not stated otherwise.

\subsection{Escape Room Experience}

The following results are reported for $39$ participants without prior escape room experience, of whom $21$ are in the reactive condition and 18 are in the proactive condition. Among participants with prior escape room experience, $7$ are in the reactive condition and $10$ are in the proactive condition.

Participants without prior escape room experience show a significantly shorter time for solving Puzzle four ($p = .017, d = 0.81$) when paired with the proactive model ($M = 445.54, SD = 180.71$) than with the reactive model ($M = 700.33, SD = 367.30$). In addition, significant differences emerge in robot perception. Participants without prior escape room experience rate the proactive model ($M = 3.20, SD = 0.58$) significantly higher in Animacy ($p = .028, d = 0.72$) than the reactive model ($M = 2.71, SD = 0.74$). Competence is also rated higher ($p = .028, d = 0.71$) for the proactive model ($M = 6.38, SD = 1.02$) than for the reactive model ($M = 5.41, SD = 1.61$) in this group. By contrast, participants with prior escape room experience rate the reactive model ($M = 4.29, SD = 0.53$) significantly higher in Likeability ($p = .034, d = 1.18$) than the proactive interaction model ($M = 3.68, SD = 0.50$).

% The Mann-Whitney U test is used to indicate significances between the interaction models. As expected, the number of interactions between the reactive interaction model ($M = 29.07, SD = 12.59$) and the proactive interaction model ($M = 97.86, SD = 21.35$) is significantly different  ($p < .001$). There are no significances between the interaction models regarding the robot perception measures for neither Godspeed and RoSAS scales nor the objective performance measures. However, the proactive condition ($71.42\%$) has a significantly ($p = .039$) lower success rate than the reactive condition ($92.86\%$).

\subsection{LLM Experience}
Based on the questionnaire, we classify $42$ participants as experienced with LLMs (reactive: $n = 19$, proactive: $n = 23$) and $14$ as inexperienced (reactive: $n = 9$, proactive: $n = 5$). Given the small sample size of participants without LLM experience, we identify only one significant difference in this group: Anthropomorphism ($p = .040, d = 1.06$) which rated higher when paired with the proactive model ($M = 2.76, SD = 0.33$) as opposed to the reactive model ($M = 2.13, SD = 0.69$).
For participants with LLM experience, we observe several significant differences in puzzle-solving time. This group shows significantly faster solving times for Puzzle one with the reactive model ($p = .036, d = 0.66; M = 775.69, SD = 199.37$) than with the proactive model ($M = 921.84, SD = 238.08$). For Puzzle two, solving time is also significantly lower with the reactive interaction model ($p < .001, d = 1.25; M = 581.45, SD = 178.30$) than with the proactive model ($M = 905.67, SD = 298.90$). By contrast, for Puzzle four, the proactive model ($M = 454.95, SD = 159.3$) results in significantly lower solving times ($p = .034, d = 0.71$) than the reactive model ($M = 665.57, SD = 368.19$) for participants with LLM experience.

\subsection{Robot Experience}
We classify $36$ participants as experienced with robots (reactive: $n = 16$, proactive: $n = 20$) and $20$ as inexperienced (reactive: $n = 12$, proactive: $n = 8$). Among participants with prior robot experience, we observe a significant difference in the time needed to solve Puzzle three ($p = .002, d = 1.42$), with faster performance in the proactive condition ($M = 247.01, SD = 75.19$) than in the reactive condition ($M = 598.71, SD = 302.98$).

\begin{figure}[t]
\vspace{5pt}
    \centering
    \begin{subfigure}{0.48\textwidth}
        \centering
        % Set a fixed height; width will scale automatically
        \includegraphics[width=\linewidth]{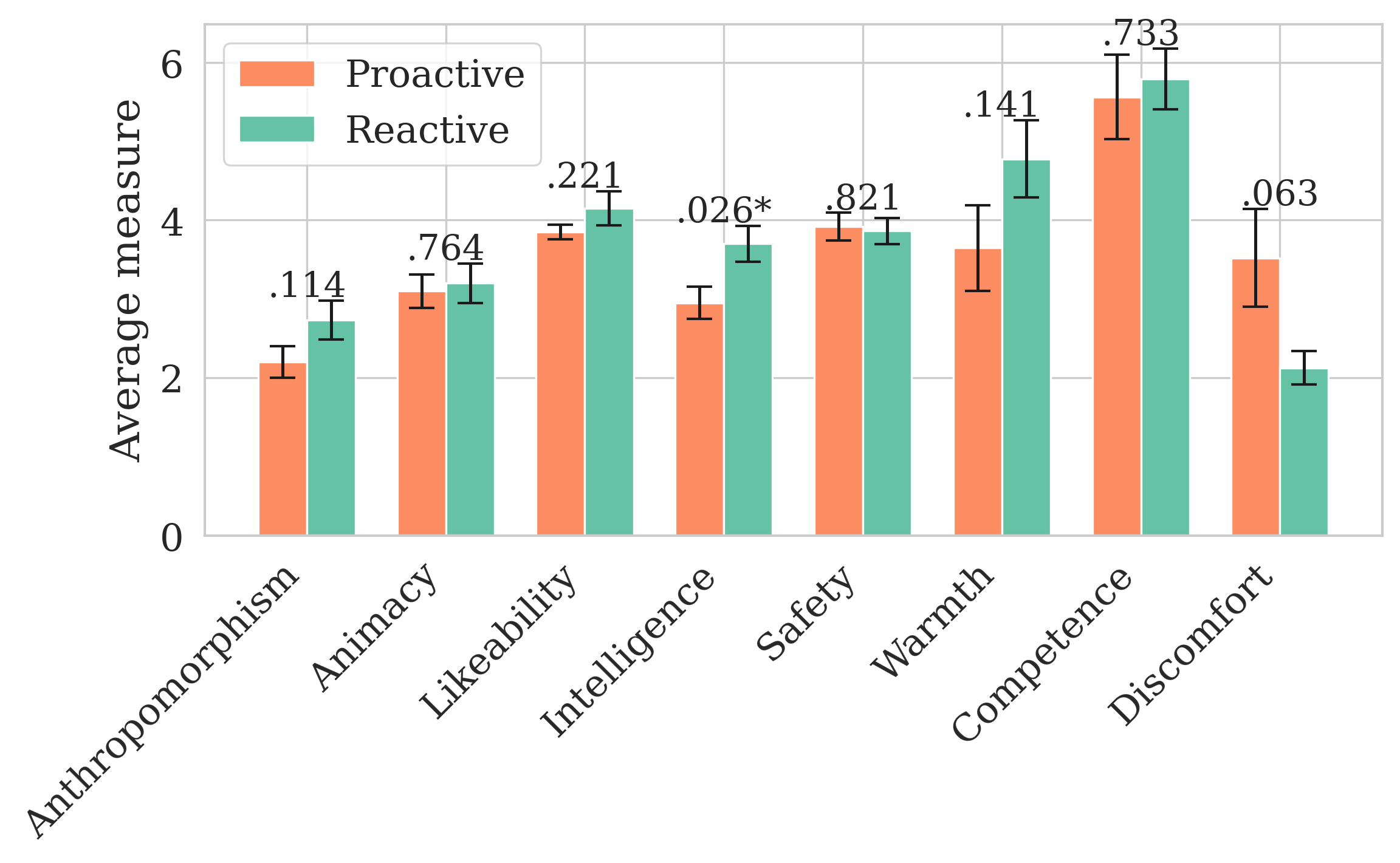}
        \caption{}
        \label{fig:experience}
    \end{subfigure}
    \hfill
    \begin{subfigure}{0.48\textwidth}
        \centering
        \includegraphics[width=\linewidth]{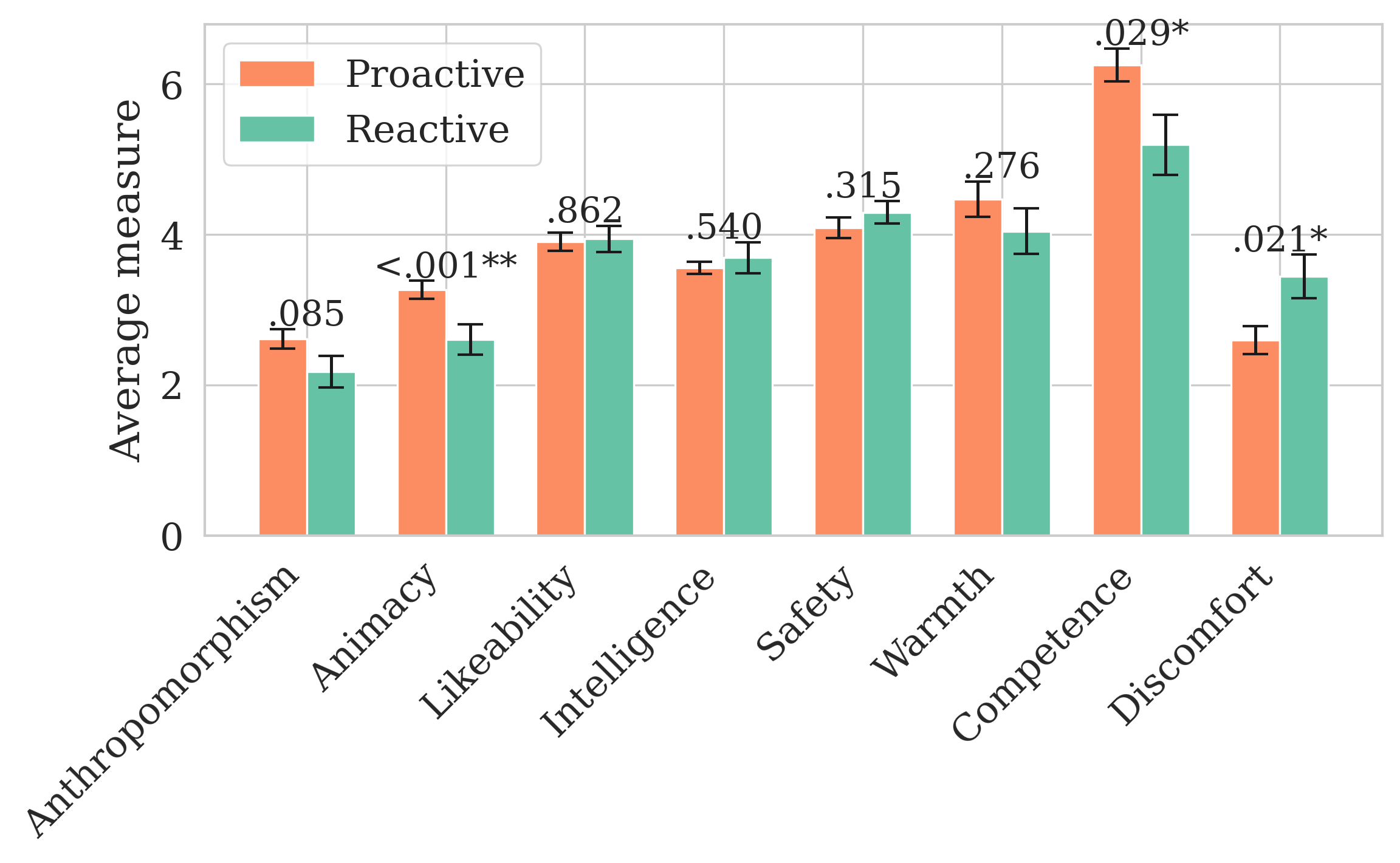}
        \caption{}
        \label{fig:not_experience}
    \end{subfigure}
    \caption{Godspeed and RoSAS scores for the proactive and reactive interaction models for participants with prior robot experience (a) and without prior robot experience (b). Reported values indicate $p$-values, *$p < .05$, **$p < .001$.}
   \label{fig:robotexe_percept}
\end{figure}

The largest differences for both groups, however, appear in robot perception. As shown in Figure \ref{fig:experience}, participants with prior robot experience rate the reactive model ($M = 3.70, SD = 0.79$) significantly higher in perceived intelligence than the proactive model ($M = 2.95, SD = 0.58$). By contrast, participants without prior robot experience report higher Competence for the proactive model ($M = 7.67, SD = 0.97$) than for the reactive model ($M = 5.19, SD = 1.61$), as illustrated in Figure \ref{fig:not_experience}. They also rate the proactive model as higher in Animacy ($M = 3.26, SD = 0.54$) than the reactive model ($M = 2.60, SD = 0.81$). In addition, participants without prior robot experience report significantly greater Discomfort with the reactive model ($M = 3.44, SD = 1.16$) than with the proactive model ($M = 2.59, SD = 0.83$). When using Student’s t-test an additional significance ($p = .023, d = 1.13$) can be identified for Discomfort for participants with robot experience, as they rate the proactive model ($M = 3.52, SD = 1.75$) higher than the reactive ($M = 2.13, SD = 0.73$).

% % Imagine the number of the hypothesis here
% To test the hypothesis \ref{seq:hypothesis}, the consistency of outcomes between the models is investigated with the Fligner-Killeen test. As shown in Figure \ref{fig:perception_variance} there are significant differences between the interaction models for the Godspeed and RoSAS measures. For Animacy, the reactive interaction model shows significantly higher variance ($p = .028$; $M = 2.86, SD = 0.88$) than the proactive model ($M = 3.21, SD = 0.55$). Similarly, for Likeability, the reactive model ($M = 4.03, SD = 0.72$) displays significantly higher variance ($p = .032$) than the proactive model ($M = 3.89, SD = 0.47$). Across both interaction models, the RoSAS measures Warmth, Competence, and Discomfort also show generally high variance, although there is no significant difference in the variance between the interaction models. The difference in variance for the time to solve a puzzle between the interaction models is illustrated in Figure \ref{fig:Timeline}, with the variance of the time required to solve Puzzle four between the reactive model ($M = 654.28, SD = 353.99$) and the proactive model ($M = 463.48, SD = 175.21$) being the only significant one.

\subsection{Extraversion}
Thirty participants are classified as extroverted based on their responses to the subset of BFI-10 items. Of these, $12$ are paired with the proactive interaction model and $18$ with the reactive model. Twenty-six participants are classified as introverted (reactive: $n = 10$, proactive: $n = 16$). Participants classified as introverted are significantly faster in solving Puzzle two when paired with the reactive model ($p = .023, d = 0.93; M = 611.13, SD = 192.63$) than when paired with the proactive model ($M = 841.12, SD = 273.53$). They are, however, significantly faster on Puzzle four with the proactive interaction model ($p = .022, d = 1.16; M = 469.0, SD = 176.89$) than with the reactive model ($M = 795.06, SD = 357.34$).

In addition, we observe significant differences in robot perception. Introverted participants rate the proactive model ($M = 6.13, SD = 1.07$) significantly higher in Competence ($p = .019, d = 1.19$) than the reactive model ($M = 4.52, SD = 1.73$). Meanwhile, extroverted participants rate the reactive model ($M = 3.83, SD = 0.68$) as significantly higher in perceived Intelligence ($p = .042, d = 0.76$) than the proactive model ($M = 3.35, SD = 0.55$). They also rate the reactive model ($M = 4.14, SD = 0.69$) significantly higher in Likeability ($p = .047, d = 0.70$) than the proactive interaction model ($M = 3.73, SD = 0.39$).

%%%%%%%%%%%%%%%%%%%%%%%%%%%%%%%%%%%%%%%%%%%%%%%%%%%%%%%%%%%%%%%%%%%%%%%%%%%%%%%%
\section{Discussion}

These results showcase various effects of robot interaction model on performance and robot perception metrics. They are notably all dependent on the studied compounding variables of robot, LLM and escape-room experience as well as extraversion, thus following our third hypothesis.

\subsubsection{LLM adapted communication}
The robot being controlled by an LLM, one would expect that LLM experience would lead to easier interactions with the robot and therefore faster information recovery and thus solving time. As expected, LLM experienced participants report a significantly faster solving time on Puzzle one and two, but only when paired with the reactive model instead of the proactive. Notably, the solving time with the reactive robot is also faster than for non-LLM experienced participants, regardless of the robot's interaction model. These observations present two striking factors. First, the increase in performance is only seen with the reactive robot, thus showing that the effect is related to the interactions with the robot. It would seem that LLM-experience leads to more efficient robot interactions, but only when said interactions are fully controlled by the participants. Second, the effect is also seen on Puzzle two, the puzzle for which the robot has notably no additional knowledge to provide. This indicates that these efficient interactions would allow not only to get easier access to the LLM knowledge, hence better performance in Puzzle one, but also to better use the deductive reasoning abilities of the LLM, the only ability capable of helping the participants in Puzzle two.
This suggests that LLM experience might translate into the adoption of a singular communication style with the robot that allows better usage of the LLM's abilities. The gain in effectiveness from this communication style would however be lost with the proactive robot, as it picks up on inter-human communication. The human-robot and robot-robot communication patterns would therefore be different and incompatible with one another, potentially complicating the acceptance of the robot in a group of humans.

Inversely, slower performances are observed for LLM-experienced participants with the reactive robot on the fourth puzzle. We however argue that this effect is caused by a selection bias. Participants with either no LLM experience or the proactive robot need more time for the first puzzles. Therefore, only the best groups of these conditions managed to reach and solve the fourth puzzle in the allowed time, thus skewing the results toward lower times.

\subsubsection{Knowledge confirmation}
The third puzzle being entirely based around identifying certain robots in the room ; one would expect robot-experienced participants to report significantly better performance on this puzzle. However, this trend is not observed in general, but instead only when said robot-experienced participants are paired with the proactive robot. Since the main difference is that the robot listens when not addressed, it therefore seems that robot experience leads the participants to try to solve the puzzle without the robot's help. The lack of performance increase with the reactive robot suggests that this strategy is generally inconclusive. Instead, the proactive model forcing the exchange of pre-existing knowledge with the robot leads to better usage of said knowledge through confirmation by the robot, and thus better performances with the proactive model.

\subsubsection{Discomfort shift}
The Discomfort rating of robot-experienced and non-robot-experienced participants displays an inversion, where the former prefer the reactive and the latter the proactive robot. Participants with no robot experience rate the proactive robot as higher in Animacy and Competence while robot-experienced participants rate the reactive robot higher in Intelligence. While Competence and Intelligence are very similar, they display variation in that RoSAS's Competence asks for responsiveness and reliability while Godspeed's Intelligence asks for responsibility and sensibility. The Intelligence of the reactive robot is therefore more aligned with pure performance on a given task, as opposed to the initiative and communication abilities highlighted by the proactive model's Competence.
These shifts in both comfort and perceived type of intelligence could hint at some form of uncanny valley effect \cite{uncanny1}. The robot-experienced participants would see the robot as less human-like and prefer therefore a reactive, more controllable robot that falls under the valley phenomenon. Meanwhile, inexperienced participants would see the robot as more human-like, as seen from the Animacy, and therefore prefer a proactive, more sentient-feeling robot standing above the valley window. %~\cite{uncanny2}.

\subsubsection{Heterogeneity preference}
Comparing robot perception metrics reveals a global trend between two subpopulations. In the context of a collaborative escape room with a robot, we distinguish participants who may be uncomfortable with the situation, participants that have no robot or escape room experience and introverted participants, opposed to participants that would be comfortable with it, participants with robot or escape room experience and extroverted participants.
Uncomfortable participants seem to generally favor the proactive robot over the reactive one. They all rate the proactive robot higher in Competence, while participants with no robot or escape room experience rate it also higher in Animacy and non-robot-experienced participants display lower discomfort with it than with the reactive robot. Participants with no LLM experience, which could also be added in this group, notably also rate the proactive robot higher in Anthropomorphism. Uncomfortable participants would therefore show appreciation for a proactive figure like the proactive robot, standing up and leading the conversation and investigation, thus appearing as more human and competent than its reactive counterpart.

In contrast, participants comfortable with the situation tend to favor the reactive robot. Extroverted and robot-experienced participants rate the reactive robot as more Intelligent, extroverted and escape-room-experienced participants rate it as more Likeable, and robot-experienced participants display lower discomfort with it. This would show an inverse effect where confident participants would prefer a more docile and following robot interaction model like the reactive one. Notably, the shift from characterizing the robot as Competent to Intelligent translates this shift in expectation, similarly to the \textit{Discomfort shift} effect.

These observations expose an effect wherein participants prefer interacting with a robot that completes their own experience. If they are comfortable with the situation, they favor a passive robot that only gives information when prompted about them. If the situation is uncomfortable, they would rather interact with a more proactive robot that will guide them and give opinions and ideas.

%%%%%%%%%%%%%%%%%%%%%%%%%%%%%%%%%%%%%%%%%%%%%%%%%%%%%%%%%%%%%%%%%%%%%%%%%%%%%%%%
\section{Conclusion}

In this work, we conducted escape room-based human-robot interaction experiments in a multi-human social setting. We compared proactive and reactive interaction models and found that both preferences and performances are strongly influenced by participants’ prior experience, both with the task and with the robot’s underlying technology, as well as personality traits like extraversion.
% First, participants with prior LLM experience appear to develop a distinct communication pattern when interacting with LLM-based systems. This adapted communication style improves performance when participants can intentionally direct it towards the robot, but not when the system also processes human--human interaction as input. Second, participants with robot experience appear to approach robots with a different set of expectations and opinions than participants without such experience. Their preferences are partly reversed: participants without robot experience perceive proactive interaction as more pleasant, whereas participants with robot experience prefer the reactive model. Lastly, in a social setting, Competence, Animacy and Likeability are all heavily influenced by the social role the robot is expected to fill in the group. Proactive groups with clear objectives and drive favor interacting with a reactive, task-specific robot. Meanwhile, indecisive groups happily welcome the addition of a more proactive, leader like robot that could give direction.
Our findings offer useful insights into the integration of robots in social settings. They suggest that current limitations of LLM-based systems may encourage users to adopt interaction patterns that differ from more natural human-human communication, thereby complicating the robot’s acceptance as an in-group social partner. Simultaneously, this observation may partly reflect the use of ChatGPT 4o, a general-purpose model designed not only for human-like interaction but also for tasks such as programming and summarization, where human-like conversational behavior is not always the primary objective. This finding would therefore benefit from further study using a specialized LLM designed explicitly for embodied human-robot interaction. Our main finding revolves around robot perception and shows that designing robots that integrate smoothly into social settings is complicated by the fact that users bring different expectations depending on their familiarity with the situation and the technology. The robot's social image is tied to the role it is expected to fill in the social group and therefore depends on the needs of said group. Preferred interaction characteristics are therefore dependent on the traits and roles already exhibited by the group, along with the circumstance-based competences and experience.

% \addtolength{\textheight}{-12cm}   % This command serves to balance the column lengths
                                  % on the last page of the document manually. It shortens
                                  % the textheight of the last page by a suitable amount.
                                  % This command does not take effect until the next page
                                  % so it should come on the page before the last. Make
                                  % sure that you do not shorten the textheight too much.

%%%%%%%%%%%%%%%%%%%%%%%%%%%%%%%%%%%%%%%%%%%%%%%%%%%%%%%%%%%%%%%%%%%%%%%%%%%%%%%%

%%%%%%%%%%%%%%%%%%%%%%%%%%%%%%%%%%%%%%%%%%%%%%%%%%%%%%%%%%%%%%%%%%%%%%%%%%%%%%%%

%%%%%%%%%%%%%%%%%%%%%%%%%%%%%%%%%%%%%%%%%%%%%%%%%%%%%%%%%%%%%%%%%%%%%%%%%%%%%%%%
% \section*{APPENDIX}

% Appendixes should appear before the acknowledgment.

% \section*{ACKNOWLEDGMENT}

% The preferred spelling of the word ÒacknowledgmentÓ in America is without an ÒeÓ after the ÒgÓ. Avoid the stilted expression, ÒOne of us (R. B. G.) thanks . . .Ó  Instead, try ÒR. B. G. thanksÓ. Put sponsor acknowledgments in the unnumbered footnote on the first page.

%%%%%%%%%%%%%%%%%%%%%%%%%%%%%%%%%%%%%%%%%%%%%%%%%%%%%%%%%%%%%%%%%%%%%%%%%%%%%%%%

\bibliography{bib}

@article{BartneckGodspeed,
  author  = {Christoph Bartneck and Dana Kuli{\'c} and Elizabeth Croft and Susana Zoghbi},
  title   = {Measurement Instruments for the Anthropomorphism, Animacy, Likeability, Perceived Intelligence, and Perceived Safety of Robots},
  journal = {International Journal of Social Robotics},
  year    = {2009},
  volume  = {1},
  pages   = {71--81},
  doi     = {10.1007/s12369-008-0001-3}
}

@incollection{BartneckGodspeedReference,
  title={Godspeed questionnaire series: Translations and usage},
  author={Bartneck, Christoph},
  booktitle={International handbook of behavioral health assessment},
  pages={1--35},
  year={2023},
  publisher={Springer}
}

@inproceedings{BeckerAttitude,
  author    = {D. Becker and K. Ahrens and C. G{\"a}de and E. Strahl and S. Wermter},
  title     = {Robots with Attitudes: Influence of {LLM}-Driven Robot Personalities on Motivation and Performance},
  year      = {2025},
  doi       = {10.1145/3765766.3765780},
  booktitle = {Proceedings of the 13th International Conference on Human-Agent Interaction},
  pages     = {40--49}
}

@inproceedings{CarpinellaRoSAS,
  author    = {Colleen M. Carpinella and Alisa B. Wyman and Michael A. Perez and Steven J. Stroessner},
  title     = {The Robotic Social Attributes Scale ({RoSAS}): Development and Validation},
  booktitle = {Proceedings of the 2017 ACM/IEEE International Conference on Human-Robot Interaction ({HRI})},
  year      = {2017},
  pages     = {254--262},
  doi       = {10.1145/2909824.3020208}
}

@inproceedings{gade2024elmira,
  author    = {C. G{\"a}de and O. {\"O}zdemir and C. Weber and S. Wermter},
  title     = {Embodying language models in robot action},
  booktitle = {Proceedings of the 32nd European Symposium on Artificial Neural Networks, Computational Intelligence and Machine Learning ({ESANN} 2024)},
  pages     = {625--630},
  year      = {2024},
  doi       = {10.14428/esann/2024.es2024-143}
}

@article{HoAlternativeToGodspeed,
  author  = {H. Chin-Chang and K. F. MacDorman},
  title   = {Revisiting the uncanny valley theory: Developing and validating an alternative to the {Godspeed} indices},
  journal = {Computers in Human Behavior},
  volume  = {26},
  number  = {6},
  pages   = {1508--1518},
  year    = {2010},
  note    = {Online Interactivity: Role of Technology in Behavior Change},
  doi     = {10.1016/j.chb.2010.05.015}
}

@misc{Multi-party,
      title={Fast Multi-Party Open-Ended Conversation with a Social Robot}, 
      author={Giulio Antonio Abbo and Maria Jose Pinto-Bernal and Martijn Catrycke and Tony Belpaeme},
      year={2025},
      eprint={2503.15496},
      archivePrefix={arXiv},
      primaryClass={cs.HC}
}

@article{NeuenswanderShortRoSAS,
  title={Measuring fundamental aspects of the social perception of robots: development and validation of a shortened version of the RoSAS (RoSAS-SF)},
  author={Neuenswander, Kelsey L and Dash, Abhinanda and Koya, Prianka D and Lin, Lin and Gillespie, Grace SR and Stroessner, Steven J},
  journal={International Journal of Social Robotics},
  volume={17},
  number={6},
  pages={1097--1112},
  year={2025}
}

@article{personalityandriella,
  author  = {A. Andriella and H. Siqueira and D. Fu and S. Magg and P. Barros and S. Wermter and C. Torras and G. Aleny{\`a}},
  title   = {Do {I} Have a Personality? Endowing Care Robots with Context-Dependent Personality Traits},
  journal = {International Journal of Social Robotics},
  volume  = {13},
  pages   = {2081--2102},
  year    = {2021},
  doi     = {10.1007/s12369-020-00690-5}
}

@article{KuchenbrandtGroupMembership,
  title={When a robot’s group membership matters: Anthropomorphization of robots as a function of social categorization},
  author={Kuchenbrandt, Dieta and Eyssel, Friederike and Bobinger, Simon and Neufeld, Maria},
  journal={International Journal of Social Robotics},
  volume={5},
  number={3},
  pages={409--417},
  year={2013},
  publisher={Springer}
}

@inproceedings{Pan2018EvaluatingSP,
  author    = {Matthew K. X. J. Pan and Elizabeth A. Croft and G{\"u}nter Niemeyer},
  title     = {Evaluating Social Perception of Human-to-Robot Handovers Using the Robot Social Attributes Scale ({RoSAS})},
  booktitle = {Proceedings of the 2018 ACM/IEEE International Conference on Human-Robot Interaction ({HRI})},
  year      = {2018},
  pages     = {443--451},
  doi       = {10.1145/3171221.3171257}
}

@inproceedings{FuExpressions,
  author    = {D. Fu and A. Fares and S. Wermter},
  title     = {The Robot in the Room: Influence of Robot Facial Expressions and Gaze on Human-Human-Robot Collaboration},
  booktitle = {2023 32nd IEEE International Conference on Robot and Human Interactive Communication ({RO-MAN})},
  year      = {2023},
  pages     = {85--91},
  doi       = {10.1109/RO-MAN57019.2023.10309334}
}

@article{DryerExtroversionIntroversionHRI,
  author  = {C. Dryer},
  title   = {Getting Personal with Computers: How to Design Personalities for Agents},
  journal = {Applied Artificial Intelligence},
  volume  = {13},
  pages   = {273--295},
  year    = {1999},
  doi     = {10.1080/088395199117423}
}

@inproceedings{LimRobotPersonality,
  author    = {M. Y. Lim and J. D. A. Lopes and D. A. Robb and B. W. Wilson and M. Moujahid and {De Pellegrin}, Emanuele and H. Hastie},
  title     = {We are all Individuals: The Role of Robot Personality and Human Traits in Trustworthy Interaction},
  booktitle = {2022 31st IEEE International Conference on Robot and Human Interactive Communication ({RO-MAN})},
  year      = {2022},
  pages     = {538--545},
  doi       = {10.1109/RO-MAN53752.2022.9900772}
}

@inproceedings{nigro2024socialgrouphumanrobotinteraction,
  title={Social group human-robot interaction: A scoping review of computational challenges},
  author={Nigro, Massimiliano and Akinrintoyo, Emmanuel and Salomons, Nicole and Spitale, Micol},
  booktitle={2025 20th ACM/IEEE International Conference on Human-Robot Interaction (HRI)},
  pages={468--478},
  year={2025},
  organization={IEEE}
}

@inproceedings{JamshadProactiveRobotInTeam,
  title={Taking initiative in human-robot action teams: How proactive robot behaviors affect teamwork},
  author={Jamshad, Rabeya and Haripriyan, Arthi and Sonti, Advika and Simkins, Susan and Riek, Laurel D},
  booktitle={Companion of the 2024 ACM/IEEE International Conference on Human-Robot Interaction},
  pages={559--562},
  year={2024}
}

@article{HoffmanCollaboration,
  author  = {G. Hoffman},
  title   = {Evaluating Fluency in Human--Robot Collaboration},
  journal = {IEEE Transactions on Human-Machine Systems},
  year    = {2019},
  volume  = {49},
  number  = {3},
  pages   = {209--218},
  doi     = {10.1109/THMS.2019.2904558}
}

@article{ahn2022icanisay,
  title={Do as i can, not as i say: Grounding language in robotic affordances},
  author={Ahn, Michael and Brohan, Anthony and Brown, Noah and Chebotar, Yevgen and Cortes, Omar and David, Byron and Finn, Chelsea and Fu, Chuyuan and Gopalakrishnan, Keerthana and Hausman, Karol and others},
  journal={arXiv preprint arXiv:2204.01691},
  year={2022}
}

@inproceedings{NgRoleplaying,
  title={Role-playing with robot characters: Increasing user engagement through narrative and gameplay agency},
  author={Ng, Spencer and Lin, Ting-Han and Li, You and Sebo, Sarah},
  booktitle={Proceedings of the 2024 ACM/IEEE International Conference on Human-Robot Interaction},
  pages={522--532},
  year={2024}
}

@article{SurveyHRIGoodrich,
  author  = {Michael A. Goodrich and Alan C. Schultz},
  title   = {Human-Robot Interaction: A Survey},
  journal = {Foundations and Trends in Human-Computer Interaction},
  year    = {2007},
  volume  = {1},
  number  = {3},
  pages   = {203--275},
  doi     = {10.1561/1100000005}
}

@inproceedings{zhang2015human,
  author    = {Zhang, Yu and Narayanan, Vignesh and Chakraborti, Tathagata and Kambhampati, Subbarao},
  title     = {A Human Factors Analysis of Proactive Support in Human-Robot Teaming},
  booktitle = {2015 IEEE/RSJ International Conference on Intelligent Robots and Systems ({IROS})},
  pages     = {3586--3593},
  year      = {2015}
}

@article{nikolaidis2017human,
  author  = {Nikolaidis, Stefanos and Hsu, David and Srinivasa, Siddhartha},
  title   = {Human-Robot Mutual Adaptation in Collaborative Tasks: Models and Experiments},
  journal = {The International Journal of Robotics Research},
  volume  = {36},
  number  = {5--7},
  pages   = {618--634},
  year    = {2017}
}

@article{dong2025happiness,
  author  = {Dong, Jiayuan and Jeon, Myounghoon},
  title   = {Happiness Improves Perceptions and Game Performance in an Escape Room, Whereas Anger Motivates Compliance with Instructions from a Robot Agent},
  journal = {International Journal of Human-Computer Studies},
  volume  = {202},
  pages   = {103547},
  year    = {2025}
}

@article{jamshad2026human,
  author  = {Jamshad, Rabeya and Haripriyan, Arthi and Sonti, Advika and Simkins, Susan and Riek, Laurel D.},
  title   = {Human-Robot Action Teams: How Robot Proactivity Affects Teamwork},
  journal = {Computers in Human Behavior: Artificial Humans},
  pages   = {100269},
  year    = {2026}
}

@inproceedings{buchem2025escape,
  author    = {Buchem, Ilona and B{\"a}cker, Niklas and Bruch, Hanna},
  title     = {Escape Room with the {NAO} Robot for {AI} Education: Comparing Cooperative and Competitive Play},
  booktitle = {International Conference on Robotics in Education ({RiE})},
  pages     = {373--379},
  year      = {2025}
}

@inproceedings{gao2019fast,
  author    = {Gao, Yuan and Sibirtseva, Elena and Castellano, Ginevra and Kragic, Danica},
  title     = {Fast Adaptation with Meta-Reinforcement Learning for Trust Modelling in Human-Robot Interaction},
  booktitle = {2019 IEEE/RSJ International Conference on Intelligent Robots and Systems ({IROS})},
  pages     = {305--312},
  year      = {2019}
}

@inproceedings{NICO,
  author    = {Matthias Kerzel and Erik Strahl and Sven Magg and Nicol{\'a}s Navarro-Guerrero and Stefan Heinrich and Stefan Wermter},
  title     = {{NICO} - Neuro-Inspired COmpanion: A Developmental Humanoid Robot Platform for Multimodal Interaction},
  booktitle = {Proceedings of the IEEE International Symposium on Robot and Human Interactive Communication ({RO-MAN})},
  year      = {2017},
  pages     = {113--120},
  doi       = {10.1109/ROMAN.2017.8172289}
}

@article{uncanny1,
  author  = {Mori, Masahiro and MacDorman, Karl F. and Kageki, Norri},
  title   = {The Uncanny Valley [From the Field]},
  journal = {IEEE Robotics \& Automation Magazine},
  year    = {2012},
  volume  = {19},
  number  = {2},
  pages   = {98--100},
  doi     = {10.1109/MRA.2012.2192811}
}

@article{uncanny2,
  author  = {Boyoung Kim and {de Visser}, Ewart and Elizabeth Phillips},
  title   = {Two Uncanny Valleys: Re-Evaluating the Uncanny Valley Across the Full Spectrum of Real-World Human-Like Robots},
  journal = {Computers in Human Behavior},
  volume  = {135},
  pages   = {107340},
  year    = {2022},
  doi     = {10.1016/j.chb.2022.107340}
}

@inproceedings{radford2023robust,
  author    = {Radford, Alec and Kim, Jong Wook and Xu, Tao and Brockman, Greg and McLeavey, Christine and Sutskever, Ilya},
  title     = {Robust Speech Recognition via Large-Scale Weak Supervision},
  booktitle = {International Conference on Machine Learning},
  pages     = {28492--28518},
  year      = {2023}
}

@article{rammstedt2007measuring,
  author  = {Rammstedt, Beatrice and John, Oliver P.},
  title   = {Measuring Personality in One Minute or Less: A 10-Item Short Version of the Big Five Inventory in English and German},
  journal = {Journal of Research in Personality},
  volume  = {41},
  number  = {1},
  pages   = {203--212},
  year    = {2007}
}

@article{TrustPriorExperience2017,
author = {Tracy L. Sanders and Keith MacArthur and William Volante and Gabriella Hancock and Thomas MacGillivray and William Shugars and P. A. Hancock},
title ={Trust and Prior Experience in Human-Robot Interaction},
journal = {Proceedings of the Human Factors and Ergonomics Society Annual Meeting},
volume = {61},
number = {1},
pages = {1809-1813},
year = {2017}
}

@article{PersonalityHRI2017,
author = {Tricia Santamaria and Dan Nathan-Roberts},
title ={Personality Measurement and Design in Human-Robot Interaction: A Systematic and Critical Review},
journal = {Proceedings of the Human Factors and Ergonomics Society Annual Meeting},
volume = {61},
number = {1},
pages = {853-857},
year = {2017}
}
\bibliographystyle{IEEEtran}

\end{document}